\documentclass{Interspeech2024}

\title{Joint Learning of Context and Feedback Embeddings in Spoken Dialogue}
\name{Livia}{Qian}
\name{Gabriel}{Skantze}

\address{
  KTH Royal Institute of Technology}
\email{liviaq@kth.se, skantze@kth.se}

\keywords{conversational systems, representation learning, unsupervised learning, backchannel, contrastive learning, feedback, dialogue, function}

\begin{document}

\maketitle

\begin{abstract}
    
Short feedback responses, such as backchannels, play an important role in spoken dialogue. So far, most of the modeling of feedback responses has focused on their timing, often neglecting how their lexical and prosodic form influence their contextual appropriateness and conversational function. In this paper, we investigate the possibility of embedding short dialogue contexts and feedback responses in the same representation space using a contrastive learning objective. In our evaluation, we primarily focus on how such embeddings can be used as a context-feedback appropriateness metric and thus for feedback response ranking in U.S. English dialogues. Our results show that the model outperforms humans given the same ranking task and that the learned embeddings carry information about the conversational function of feedback responses. %Through our experiments, we find that there is substantial information in jointly training short contexts and feedbacks.

\end{abstract}

\section{Introduction}

%Spoken dialogue systems engaging in direct communication with humans warrant the need for reliable speech representations. 
In spoken communication, the role of feedback is crucial to streamline conversations and convey the interlocutor's state-of-mind, e.g., to signal attention, understanding, or attitude~\cite{Schegloff,clark1996using, buschmeier2012adapting}. Feedback can be expressed in different forms through voice, face and body~\cite{poggi2007mind, bevacqua2010multimodal, morency2010probabilistic}. Vocal backchannels like `uh-huh', `wow' and `yeah' often serve the role of non-interrupting feedback in the background that the active speaker may or may not explicitly react to~\cite{Yngve70, 10.1093/applin/19.2.204}. They constitute a subset of short vocal feedback responses including yes/no responses and agreements, which may or may not be regarded as a separate turn~\cite{10.1093/jos/9.1.1, WARD20001177}.

%Like dialogue acts, feedback responses can be categorized into various communicative functions~\cite{10.1093/jos/9.1.1, mumin}. By defining possible sets of feedback functions, recent work highlights their diverse roles in conversations and the context-dependent nature of their functions~\cite{boudin2021multimodal, figueroa-etal-2023-classification}. Other studies suggest that speakers prosodically align their feedback with the preceding speech from the interlocutor, further emphasizing their context-dependence~\cite{heldner2013backchannel, figueroa:hal-04192125}.
%Recent work defining a possible set of feedback functions show the diversity of feedback in terms of their role in a conversation, as well the dependence of context in terms of how their function should understood~\cite{boudin2021multimodal, figueroa-etal-2023-classification}.

Like dialogue acts, feedback responses can be classified into various communicative functions~\cite{10.1093/jos/9.1.1, mumin, DBLP:conf/interspeech/BuschmeierMWKW11}. Recent work defines different types of feedback functions, emphasizing their diverse roles and context-dependent nature in conversations~\cite{boudin2021multimodal, figueroa-etal-2023-classification}. Studies also suggest that speakers align their feedback prosodically with the preceding speech from the interlocutor, further underlining their context-dependence~\cite{heldner2013backchannel, figueroa:hal-04192125}.

As feedback is crucial in human-human communication, this should also be the case for spoken dialogue systems (SDS). However, current trends in text-to-speech (TTS) and automatic speech recognition (ASR) rely mainly on modeling the main channels of speech, often ignoring the presence of backchannels and human-like conversations in general~\cite{Tan2021ASO, liesenfeld-etal-2023-timing}. In the case of generating feedback responses, most work has focused on timing, i.e., detecting suitable places in the user's speech for generation~\cite{ruede2019yeah, 10.1145/3472306.3478360, DBLP:conf/icassp/OrtegaLV20}, often ignoring the form, i.e., \textit{how} they should be generated to match the dialogue context and to convey appropriate communicative functions. 

%While it is possible to manually label different feedback functions with an annotation scheme, it is more compelling to learn the context-dependence and functional representations in an unsupervised manner. We investigate an approach that jointly embeds feedback responses and their context in the same representation space, with the contrastive learning objective shown in Figure~\ref{fig:contrastive}. We expect the learned representations to capture the functional and contextual aspects of feedback responses. 

\begin{figure}[t]
  \centering
  \includegraphics[width=\linewidth]{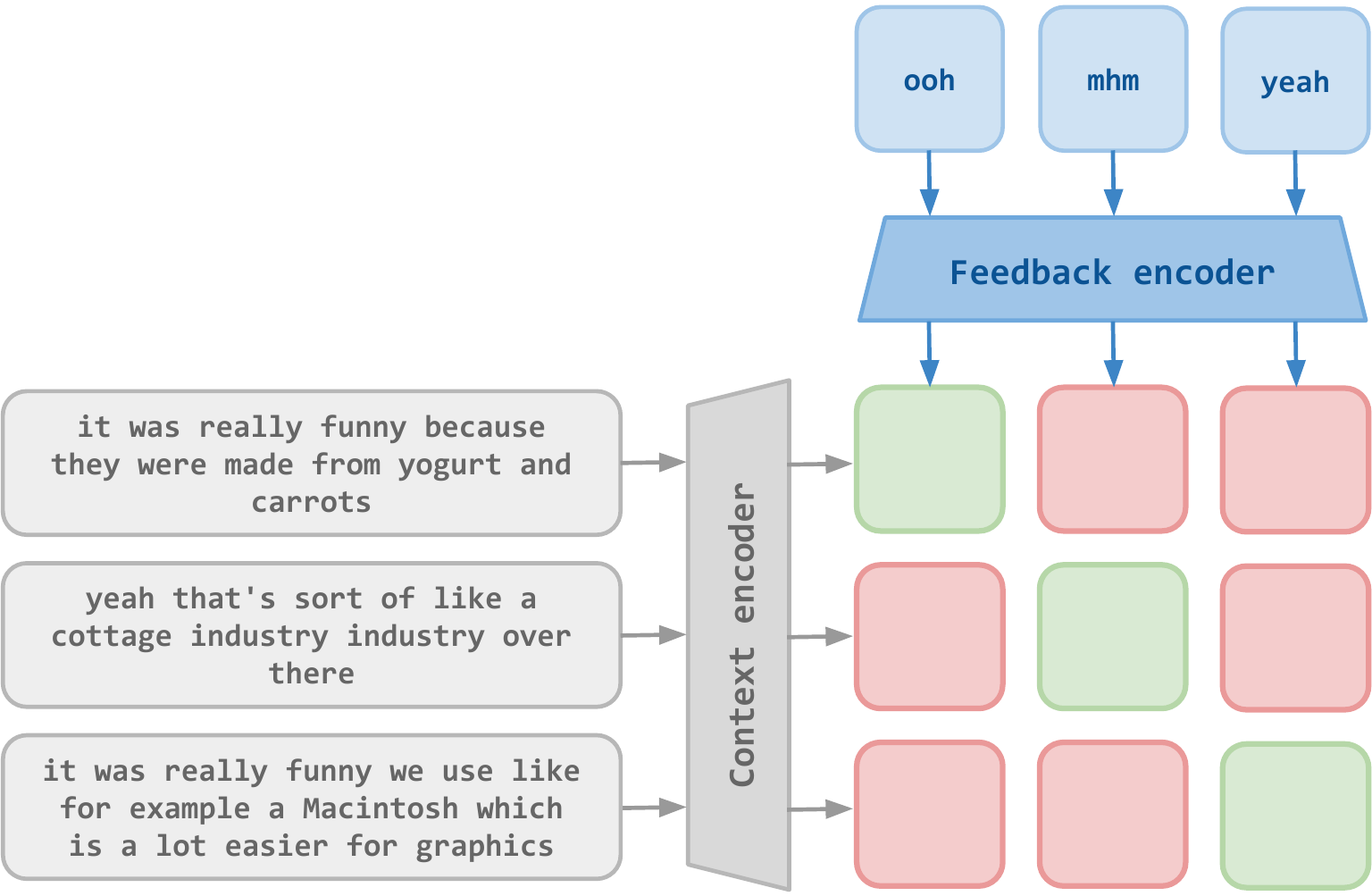}
  \caption{Outline of our contrastive learning approach where the matrix represents the similarity scores between combinations of context-feedback pairs. The green and red boxes represent positive and negative pairs, respectively. The contexts and feedback responses can be present in audio and/or text format.}
  \label{fig:contrastive}
\end{figure}

Instead of manually labeling various feedback functions with an annotation scheme, we explore an unsupervised approach to learn functional representations and encode context-dependence. Our method jointly embeds feedback responses and context in the same space using a contrastive learning objective (Figure~\ref{fig:contrastive}). We expect that the learned representations will effectively capture both the functional and contextual aspects of feedback responses.

%The general approach we adopt is illustrated in Figure~\ref{fig:contrastive}. We take short feedback responses (single isolated words of 200-1200 ms in length) along with the immediate preceding context (4000 ms). The feedback responses and contexts are then embedded separately using an audio/text encoder and projected into an N-dimensional vector. Through contrastive learning, the embeddings of correct pairs are pushed together, while the embeddings of incorrect pairs (in the same batch) and pushed apart. 

This work aligns with the paradigm of self-supervised learning using contrastive methods, as seen in SimCLR (images)~\cite{chen2020simple}, CPC (speech)~\cite{cpc} and CLIP (image/text)~\cite{radford2021learning}. Our work is also related to response ranking in dialogue systems~\cite{AlRfou2016ConversationalCC, henderson-etal-2020-convert}, which involves learning a context-response similarity score for ranking potential responses in a dialogue context. Additionally, our work shares a connection with~\cite{liesenfeld22_interspeech}, where vocal feedback responses from various languages are analyzed via UMAP clustering of acoustic properties, though their approach does not incorporate the feedback context.

% , a concept that can be applied to rank feedback responses in a dialogue context
%where a context-response similarity score is learned and then used to rank potential responses. Similarly, we could use this approach to rank potential feedback responses in a given dialogue context. Another related work is~\cite{liesenfeld22_interspeech}, where vocal feedback responses from different languages are analyzed through the UMAP clustering of acoustic properties. This approach, however, does not incorporate the feedback context.

%https://arxiv.org/abs/1911.03688
%https://arxiv.org/abs/1606.00372

% Some works have placed focus on feedback tokens [source], but even when the prosody is natural, they are good only in isolation and not in the actual context [source]. TODO continue

Recent work has shown the importance of non-lexical information (e.g., prosody) in spoken dialogue response ranking~
\cite{wallbridge21_interspeech, nath22_speechprosody}. 
Since feedback responses are short and have little lexical content, we can expect prosody to play a prominent role. Thus, we explore both acoustic and text-based approaches. 

Our main contribution in this paper is to show the potential of jointly training dialogue context and feedback responses to get representations that express the relationship between them, based on both lexical and non-lexical information. We also set up a human baseline for ranking potential feedback responses given a context. Finally, using manual labels of the conversational function of feedback responses, we show that the learned representations of both context and feedback carry information about these functions. 

\section{Method}

\subsection{Dataset}
\label{sec:dataset}
%We compared Switchboard~\cite{switchboard} and Fisher Part 1~\cite{cieri-etal-2004-fisher}, both consisting of telephone calls between native speakers of U.S. English. In both cases, the conversations were recorded on two separate channels, one for each speaker. Time-aligned transcripts are also available in both cases. Although the two corpora are similar, the main advantage of using Fisher is its considerable size ($\approx 1000$ hours). The advantage of Switchboard is that part of the data has been manually annotated based on feedback functions~\cite{figueroa-etal-2022-annotation}, which helps us to analyze the learned embeddings. 

We compared two corpora, Switchboard~\cite{switchboard} and Fisher Part 1~\cite{cieri-etal-2004-fisher}, both comprising telephone calls between native speakers of U.S. English, recorded on separate channels for each speaker, and with time-aligned transcripts available. While the two corpora share similarities, Fisher offers a significant advantage due to its substantial size (approximately 1000 hours). On the other hand, parts of Switchboard have manual annotation of feedback functions~\cite{figueroa-etal-2022-annotation}, facilitating the analysis of learned embeddings. This annotation scheme contains 10 classes: `Continue' (C), `Agree' (A), `Sympathy' (S), `Non-understanding' (U), `Disagree' (D), `Yes' (Y), `No' (N), `Disapproval' (Ds), `Mild Surprise' (MS) and `Strong Surprise' (SS).

For Switchboard, we used lists of feedback instances identified by~\cite{figueroa-etal-2023-classification}. These instances are a combination of manually labeled (as mentioned earlier) and automatically labeled data (for the majority of the dataset). To mitigate potential issues like crosstalk and ensure the effectiveness of the contrastive learning objective, we selected instances without crosstalk (verified by Switchboard annotations) or where the same speaker remained silent in the preceding 4 seconds. This ensures that the feedback speaker is not audible in the context, avoiding reliance on speaker identification during training. Additionally, feedback instances shorter than \SI{200}{\milli\second} were excluded.

The manually labeled data, excluding conversations present in the automatically labeled data, served as test data. A 90/10\% split of the automatically labeled data was used for training and validation.

For Fisher, we first extracted feedback instances by identifying a set of lexical tokens in the transcripts. We used the tokens defined in~\cite{figueroa-etal-2023-classification}, with a few additions based on different spellings. From these, we selected instances with a duration of at least \SI{200}{\milli\second}, and where the same speaker was silent before the feedback for at least 4 seconds (for reasons mentioned earlier) and after it for at least 1 second (to ensure separation from other types of speech). The transcript was missing in some cases; these samples were taken away when text embeddings were considered. Here we used a ~80/10/10\% split for training, validation and test, respectively~\footnote{The ratios are approximate as the splits have to be disjoint with respect to speakers and conversations.}.

The details of the resulting datasets are shown in Table~\ref{tab:datasets}.

% current version of the table
    
\begin{table}[th]
\caption{The details of the two datasets. Each sample consists of a context-feedback pair. %In the case of Switchboard, \textbf{m} represents the manually and \textbf{a} the automatically labeled set from~\cite{figueroa-etal-2023-classification}.
}
    \label{tab:datasets}
    \centering
    \begin{tabular}{lcc}
    \toprule
    \textbf{Setup} & \textbf{Switchboard} & \textbf{Fisher} \\
    \midrule
    \# training samples & 44261 & 66991 \\
    \# validation samples & 4910 & 8369 \\
    \# test samples & 1129 & 8375 \\
    %\# speakers & 287 (m), 4347 (a) & 11261 \\
    %\# distinct fb. tokens & 50 (m), 83 (a) & 47 \\
    %\# average fb. length & 385 (m), 422 (a) ms & 470 ms \\
    %\midrule
    %\# distinct fb. functions & 10 (+ 1) & - \\
    \bottomrule
    \end{tabular}
    \end{table}

%The preprocessing consists of loading the audio samples and resampling them to 16 kHz. We precomputed the embeddings, i.e., encoded the audio files and/or transcripts with the respective audio and/or text encoders and saved them as inputs to the model heads. This step is equivalent to freezing the pretrained models and helps avoid computational bottlenecks.

\subsection{Model setup}

We approach context and feedback modeling by joint contrastive training in the style of CLIP~\cite{radford2021learning}, as illustrated in Figure~\ref{fig:contrastive}. %We tried the CLIP and the InfoNCE~\cite{cpc} loss, with the initial results consistently favoring the former. 
Feedback responses, along with the immediate preceding context (\SI{4000}{\milli\second}), are embedded separately using an audio and/or text encoder and projected into an $M$-dimensional vector. In a batch of $N$ context-feedback pairs, the cosine similarity between all $N \times N$ combinations is computed, and the symmetric InfoNCE \cite{cpc} loss over these similarity scores is used. Through this learning objective, the embeddings of matching pairs are pushed together and those of incorrect pairs apart.

%Previous work \cite{10023234} has shown that HuBERT~\cite{hubert} embeddings contain prosodic information. Therefore, we used HuBERT but also Whisper's encoder~\cite{whisper} for encoding audio. For encoding text, we used BERT~\cite{bert}, SimCSE~\cite{gao2021simcse} and GTE~\cite{gte}~\footnote{\texttt{facebook/hubert-large-ls960-ft}, \texttt{openai/ whisper-medium}, \texttt{google-bert/ bert-large-uncased}, \texttt{princeton-nlp/ sup-simcse-bert-large-uncased}, \texttt{thenlper/gte-large}}. 

%An outline of the contrastive learning approach can be seen in Figure~\ref{fig:contrastive}.% and the loss formula is shown in Equation~\ref{ref:loss}.
%The model architecture consists of the following parts: an audio encoder (either with a text encoder or without) for encoding the context and another one for encoding the corresponding feedback. These serve the role of the backbones of our models. 

Previous work \cite{10023234} has shown that HuBERT~\cite{hubert} embeddings contain prosodic information. Therefore, we used HuBERT (large) but also Whisper's encoder (medium)~\cite{whisper} for encoding audio.
For encoding text, we used BERT (large)~\cite{bert}, SimCSE (large)~\cite{gao2021simcse} and GTE (large)~\cite{gte}, all with pre-trained weights from Huggingface~\cite{huggingface}.
Instead of fine-tuning, we extracted encodings from all layers of the respective encoders and applied weighted average pooling on them, as well as mean pooling along the sequence length (similarly to \cite{10023234}). Audio and text embeddings are concatenated when used together. The model head, either a linear layer or MLP, projected the resulting embeddings into the same space. Throughout our experiments, the feedback and context embedding models had the same configuration but were trained with separate parameters.

We compared models with different configurations and hyperparameters such as combinations of audio and text embeddings (HuBERT/Whisper + BERT/SimCSE/GTE), model head (linear layer and MLP), hidden size and/or projection size (a combination of 128, 256, 512 and 1024 in descending order across the layers), loss temperature (between 0.001 and 0.5), temperature training (true and false), batch size ($2^n$, $n$ from 8 to 13), learning rate (between 1e-4 and 1e-1) and optimizer (Adam and AdamW). We used grid search and uniform sampling for hyperparameter tuning, as well as early stopping on the validation sets to identify the best models, whose performance is subsequently reported on the test sets.

We used 8 NVIDIA GeForce RTX 3090 GPUs for all purposes. The average runtime of the hyperparameter search was 45 minutes per run for Switchboard and 70 for Fisher. The number of parameters is $\approx 3.1$M and $\approx 1$M in unimodal models with two-layer MLPs and linear layers, respectively; the same numbers are $\approx 5.2$M and $\approx2.1$M for audio-and-text models.

%The loss looks as follows:
%\begin{equation} \label{ref:loss}
%\text{loss} = \frac{1}{2} \big(CEL(\text{logits, labels}) + CEL(\text{logits.T, %labels})\big)
%\end{equation}
%where $CEL$ is the cross-entropy loss, $logits$ is an $N \times N$ matrix representing the temperature-scaled cosine similarity between every combination of $N$ context-feedback pairs, and $labels$ are the integers ranging from $1$ to $N$ (inclusive). This maximizes the cosine similarity between correct pairings (by emphasizing the diagonal of $logits$) and minimizes that of the $N^2 - N$ incorrect pairings.

\subsection{Metrics}

In this work, we primarily assess the models' capacity to rank the correct feedback instance within a set of distractors (a `batch') based on any given context, by computing and sorting by the cosine similarity between the context embedding and each feedback embedding. This aligns with the training objective, providing an overall indication of how effectively the model jointly embeds the context and feedback response. It also offers insight into the model's suitability as a feedback response ranker in a SDS. It is important to note that in ranking responses, various alternatives may be suitable, particularly for continuers like `mhm' or `yeah', which are often interchangeable. Meanwhile, feedback responses like those expressing strong surprise or disagreement may be more context-sensitive.

%As a ranking metric, we calculated the top-k accuracy where k represents the top 1\%, 10\%, 25\% and 50\% of the batch size, i.e., of the number of feedback responses considered for each context. The number of samples these fractions represent depends on the batch size; we used the same one for training, validation and test within a single model configuration, unless the validation or test set is smaller than the batch size in question, in which case the entire set is ranked. In terms of top-k performance, our main point of reference for validation is top-25\% as it corresponds to the baseline our human subjects received to solve the same task, namely to rank four feedback candidates given a context (Section \ref{sec:human}).

As a ranking metric, we calculated the top-k accuracy with k = 1\%, 10\%, 25\% and 50\%, where each of these fractions corresponds to the top k (percent) of the number of rankable feedback responses for each context (i.e., `batch size'). The batch size determines the exact number of samples to rank. We maintained a consistent batch size for training, validation, and test sets within a model configuration, unless a set is smaller, in which case the entire set is ranked. Our primary reference for validation is top-25\%, aligning with the baseline provided to human subjects in a similar task, i.e., ranking four feedback candidates (Section \ref{sec:human}).

%As a ranking metric, we calculated the top-k accuracy where k represents the top 1\%, 10\%, 25\% and 50\% of the number of feedback responses to rank for each context (i.e., 'batch size'). The batch size determines the number of samples these fractions represent. We maintained a consistent batch size for training, validation, and test sets within a model configuration, unless a set is smaller, in which case the entire set is ranked. Our primary reference for validation is top-25\%, aligning with the baseline provided to human subjects in a similar task, i.e., ranking four feedback candidates (Section \ref{sec:human}).

%As a ranking metric, we calculated the top-k accuracy where k represents the top 1\%, 10\%, 25\% and 50\% of the batch size, i.e., of the number of feedback responses considered for each context. The batch size determines the number of samples these fractions represent. We maintained a consistent batch size for training, validation, and test sets within a model configuration, unless a set is smaller, in which case the entire set is ranked. Our primary reference for validation is top-25\%, aligning with the baseline provided to human subjects in a similar task, i.e., ranking four feedback candidates (Section \ref{sec:human}).

\section{Results}

\subsection{Model results}

In Table \ref{tab:model1}, we show the results for model performance on the Switchboard and Fisher test sets (\% for top-k accuracy). Accuracy is based on the ranking of the ground truth, i.e., the true feedback response belonging to each context, and whether it is in the top k (\%). 
%We tested multiple configurations, with a special focus on the combinations of backbone models. 
We include text-only and random baselines for comparison, but these are not used in our further analyses.

\begin{table}[th]
\caption{Model performance on the Switchboard (upper) and Fisher (lower) test sets wrt. accuracy (\%). We show the best embedding combinations for audio, text, and audio + text (A+T).}
    \label{tab:model1}
    \centering
    \begin{tabular}{lcccc}
    \toprule
    \textbf{Model on Switchboard} & \textbf{t1\%} & \textbf{t10\%} & \textbf{t25\%} & \textbf{t50\%} \\ \midrule
    Audio only (HuBERT) & 3.72 & 23.83 & 45.79 & 72.19 \\
    A+T (Whisper + GTE) & \textbf{4.69} & \textbf{30.82} & \textbf{53.85} & \textbf{78.03} \\
    \midrule
    Text only (GTE) & 2.92 & 22.94 & 45.00 & 70.59 \\

    \bottomrule
    \toprule

    \textbf{Model on Fisher} & & & & \\ \midrule
    Audio only (Whisper) & 6.05 & 33.4 & 57.42 & 80.99  \\
    A+T (HuBERT + GTE) & \textbf{7.19} & \textbf{36.45} & \textbf{60.94} & \textbf{82.65} \\
    \midrule
    Text only (GTE) & 3.56 & 19.55 & 39.5 & 65.87 \\
    \bottomrule
    \toprule
    Random baseline & 1 & 10 & 25 & 50 \\
    \bottomrule
    \end{tabular}
\end{table}

After hyperparameter tuning, we noticed that the best models preferred a batch size of 4096 and one- or two-layer MLPs with hidden sizes 512, and 1024 and 512, respectively (with a final embedding size of $M=512$). No specific pattern was found in other hyperparameters. From this point on, we choose HuBERT as our best audio-only model and Whisper + GTE as our best audio-and-text model for Switchboard.

It can be seen that Switchboard is generally harder to learn, which may be due to the quality of the recordings, the dataset size, and the difference between the feedback responses of the training and test set (automatic and manual extraction). The audio-and-text embeddings consistently outperform audio, which in turn outperform text (this difference is more pronounced for Fisher). Clearly, the text embeddings in and by themselves do not contain enough information, which is expected given the limited lexical content in feedback responses. This is also in line with previous research showing the importance of prosody in spoken dialogue response ranking \cite{wallbridge21_interspeech}.% to distinguish between certain sets of feedback tokens. This is expected, considering the limited number of lexical forms present in the data.

%In Table [], we also show the sensitivity of intra- and inter-class predictions of our models. todo

\subsection{Human evaluation}
\label{sec:human}

%Since it is hard to know what ranking performance could be expected of these models, we also did a human evaluation. We curated a test set of 240 samples taken from the Switchboard test set. These samples were picked to balance the conversational functions of the feedback responses, using the manual function labels, resulting in 24 samples for each function. Using the Prolific platform, we conducted experiments where participants had to rate four feedback responses based on the extent to which they can be a possible continuation of a given context (4000 ms). The participants were asked to 1) choose one of four feedback responses that is considered the best match and 2) rate each option from a scale of 1 (worst) to 4 (best). Due to the bad quality of the recordings, we also asked about the intelligibility of the contexts.

To put the ranking performance of these models in context, a human evaluation was conducted. A test set of 240 samples was curated from the Switchboard test set, balancing the manually annotated conversational functions of the feedback responses from ~\cite{figueroa-etal-2022-annotation} (10 functions in total, resulting in 24 samples per function). Participants, recruited on the Prolific platform, rated four feedback responses based on their suitability as a continuation of a given context (\SI{4000}{\milli\second}). They were tasked to 1) choose the best match from the four responses and 2) rate each option on a scale from 1 (worst) to 4 (best) in terms of how well they match the context. Additionally, participants were asked to provide feedback on the intelligibility of the contexts due to the poor quality of the recordings.

We tried two different conditions where participants were given either 1) only audio or 2) a combination of text and audio. For each condition, 60 participants were recruited. Each participant evaluated 20 combinations, consisting of a single context and four feedback responses (one correct and three randomly sampled distractors). Additionally, as an intra-group condition, we explored the impact of ranking feedback responses with different versus the same function labels (also based on the manual annotation). Hence, half of the combinations had feedback options of the same type, and the other half had different types. Participants were limited to those indicating English as their first and primary language, and who have spent most of their first 18 years in the United States and currently reside there.

%60 participants were recruited for each of the two conditions. Each participant received 20 combinations of a single context and four feedbacks (one correct and three randomly sampled distractors). As a secondary (within-group) condition, we also explored the difference between ranking feedback responses with different vs. the same function labels (according to the manual annotation). Thus, half of the 20 combinations the participants received had feedback options of the same type and half of them had different types. The participant pool was restricted to individuals who indicated that their first and primary language is English, and who had spent most of their first 18 years living in the US and were currently located there.

The exact same tasks were given to the best models (based on HuBERT and Whisper + GTE) for comparison. 
The results are shown in Table \ref{tab:participant-performance}. As can be seen, the subjects were fairly good at ranking feedback responses with different functions (which may also indicate a greater lexical dissimilarity between function types) but fared worse when they ranked feedback responses with the same function. For the audio-only condition, the model had a similar performance as humans. When adding transcripts, the performance of humans did not improve (but they helped to increase intelligibility), while the models clearly improved and outperformed humans for this condition.

\begin{table}[th]
\caption{Model and participant performance on the curated Switchboard test set. Accuracy is based on what was selected as the best match from the feedback received (\%). Intelligibility is based on the perceived intelligibility of the contexts (\%).}
    \label{tab:participant-performance}
    \centering
    \begin{tabular}{lccc}
    \toprule
    \textbf{Setup} & \textbf{model} & \textbf{human} & \textbf{intelligibility} \\
    \midrule
    Audio, same fun. & 30.23 & 30.90 & 73.26 \\
    Audio, different fun. & 37.63 & 41.80 & 73.08 \\
    \midrule
    Audio + text, s. fun. & 40 & 28.60 & 77.69 \\
    Audio + text, d. fun. & 48.74 & 42.52 & 78.32 \\
    \midrule
    Random baseline & 25 & 25 & N/A \\
    \bottomrule
    \end{tabular}
    \end{table}

%We conducted the same experiment on random samples from the Fisher test set. todo

\begin{figure}[t]
  \centering
  \includegraphics[width=0.8\linewidth]{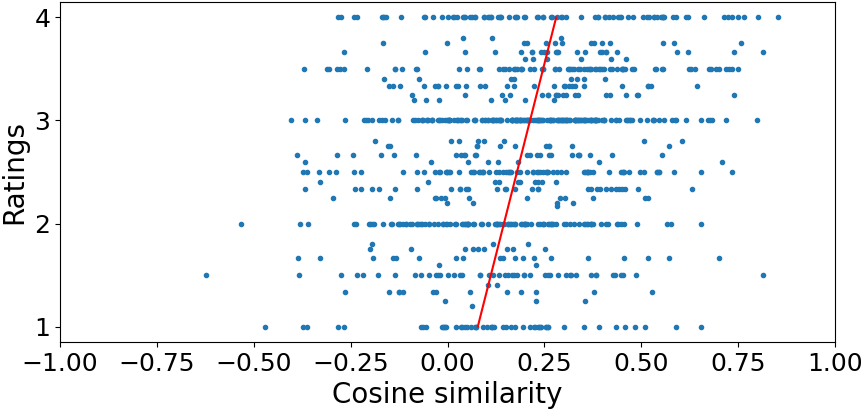}
  \caption{Scatter plot of the cosine similarity score (x-axis) assigned by our best audio-based model trained on Switchboard (HuBERT) and the participants' ratings (y-axis). For each data point, the similarity score and rating are based on context-feedback pairs with the same function label.}
  \label{fig:correlation_audio}
\end{figure}

In Figure~\ref{fig:correlation_audio}, the results of the appropriateness ratings (1-4) by humans and the cosine similarities by our top-performing audio-based model (HuBERT) are plotted. The plot corresponds to the setup of the first row in Table~\ref{tab:participant-performance}.

The model and participants had the same task -- assigning compatibility scores to the same combinations of contexts and feedback responses, which in this case are audio-only context-feedback sets with identical functions within each set. Mean scores were computed for context-feedback combinations with multiple annotator scores. There is a modest correlation between the human- and model-assigned scores ($r=0.23$). We found a similar (but weaker) correlation for context-feedback combinations with different functions ($r=0.14$).

For our best audio-and-text model (Whisper + GTE), the correlations were also positive, both for identical functions ($r=0.16$) and for different functions ($r=0.19$). All correlations were statistically significant ($p< 0.001$).

\subsection{Probing classifiers}
\label{sec:probing}

To assess if the learned representations encode information about conversational functions, we trained a probing classifier on the output embeddings produced by our best models, using the 10 types of manual function labels of the Switchboard test set~\cite{figueroa-etal-2022-annotation} as targets. For comparison, we performed the same procedure on embeddings from corresponding pre-trained encoders (HuBERT, as well as Whisper + GTE concatenated). We applied 10-fold cross-validation with an SVM\footnote{\href{https://scikit-learn.org/stable/modules/generated/sklearn.svm.SVC.html}{\texttt{sklearn.svm.SVC}}} on top (linear, C = 1). As input, we used the feedback or context embeddings alone, or their concatenation. The results are shown in Table~\ref{tab:probe}. 

\begin{table}[th]
\caption{Function classification accuracy (\%) with linear SVM, using audio or audio and text (A+T) embeddings.}
    \label{tab:probe}
    \centering
    \begin{tabular}{lccc}
    \toprule
    \textbf{Embeddings} & \textbf{fb.} & \textbf{cont.} & \textbf{fb. + cont.} \\ \midrule
    Audio (HuBERT) & 58.49 & 43.06 & 63.94  \\
    Audio (Our model) & 59.76 & 49.23 & 62.39 \\ \midrule
    A+T (Whisper + GTE) & 67.57 & 42.69 & 70.94  \\
    A+T (Our model) & 68.03 & 51.59 & 73.02 \\
    \bottomrule
    \end{tabular}
\end{table}

The learned feedback embeddings provide limited additional information about communicative functions compared to the embeddings of the pre-trained models, considering the rather small gap between their performance. In contrast, the learned context embeddings contain more information. This could be valuable for predicting suitable feedback functions following specific contexts. The superior performance of the combined feedback and context embeddings is unsurprising, considering that the function labels of the feedback responses were annotated with both context and feedback in mind \cite{figueroa-etal-2022-annotation}.

\subsection{Visualization of embedding space}

\begin{figure}[t]
  \centering
  \includegraphics[width=\linewidth]{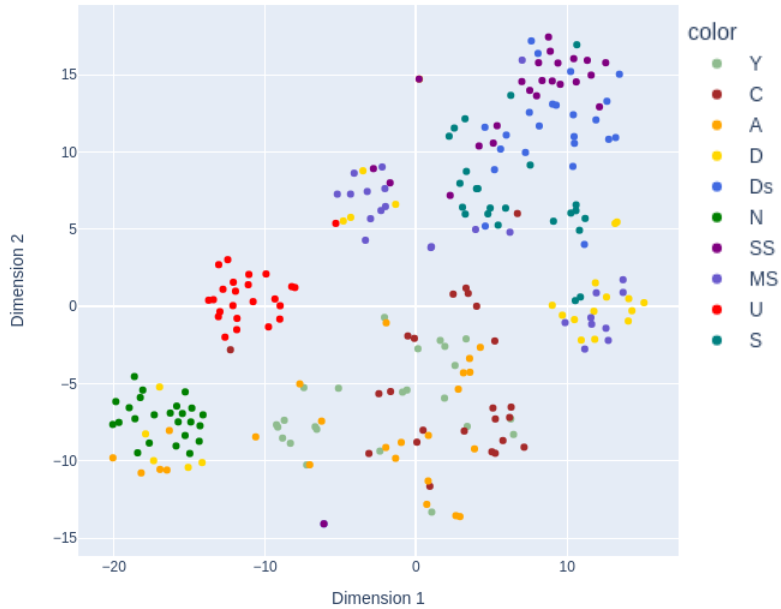}
  \caption{The t-SNE plot of learned audio-and-text embeddings where the feedback and context embeddings are concatenated. Abbreviations of functions are defined in Section~\ref{sec:dataset}. }
  \label{fig:tsne}
\end{figure}

To illustrate the learned embedding space, 
%we decided to visualize the embedding space and how the curated Switchboard test set (with balanced function labels) is encoded by this model. 
the curated Switchboard test set (with balanced function labels) is first encoded using the top-performing audio-and-text model. We then apply t-SNE on the concatenated context and feedback embeddings.

%and \ref{fig:tsne_feedback}. 
As can be seen in Figure~\ref{fig:tsne}, there is a clear clustering of different feedback functions, although some groups are merged. For example, the embeddings of `Non-understanding' (U) context-feedback pairs are neatly clustered together (e.g., `what?', `pardon?'). `Disapproval' (Ds) and `Strong Surprise' (SS) also tend to co-locate; they both express strong emotions with lexically and semantically similar elements (e.g., `ooh', `gosh' and `jeez'). The lexical token `no' can be a `No' (N) answer to a yes-no question, as well as express both `Agreement' (A) and `Disagreement' (D); some of the elements from the latter two are centered around the first group. Similarly, `Yes' (Y), `Agreement' (A) and
`Continue' (C) form their own -- although rather scattered -- cluster (examples include `yes', `mhm' and `right'). `Sympathy' (S) and `Mild Surprise' (MS) are close to  `Disapproval' (Ds), but they also form their own groups.

\section{Conclusion and Discussion}

In this paper, we explored the potential of jointly training embeddings of short feedback responses and of their immediate contexts in an unsupervised manner. The results are promising: in the task of feedback response ranking (using the cosine similarity of embeddings as a similarity metric), the model performs comparably to humans with audio alone and outperforms humans when using both audio and text. This bodes well for generating context-appropriate feedback in spoken dialogue systems. This could be achieved through direct response ranking of synthesized feedback candidates or via classification of appropriate feedback functions, guiding the synthesis process.

%show that it is possible to solve the task of feedback selection using light-weight contrastive learning, and it is possible to outperform humans on this very task. Since our models can produce similarity scores based-on cosine similarity, a possible application is feedback ranking given a context and a set of feedbacks. TODO continue

%Acousic information seems to be more important than lexical information, which is in line with the findings of~\cite{DBLP:conf/icassp/OrtegaLV20}.

%In this work, we analyzed the relationship between feedbacks and past context to simulate the simplest form of what the speech processor of an online conversational system has access to at any given point of time. The future context can also play an important role~\cite{nath22_speechprosody}, especially in the form of the interlocutor's upcoming turn. This information, however, is usually not available to the speaker, which is the suspected role of a conversational agent in this situation.

Our experiments are limited to U.S. English, but given the unsupervised nature of our method, its application to other languages is feasible. One potential avenue is cross-lingual analyses of feedback akin to~\cite{liesenfeld22_interspeech}. We hope that our work will be a step towards a unified, language-agnostic approach to modeling response tokens.

%\subsection{Limitations and future work}

Our current work shows the importance of the immediate context in determining the upcoming feedback but does not account for longer contexts 
 (e.g., utterances, turns, and previous feedback). Moreover, we do not include potentially meaningful aspects like interlocutor personality, voice characteristics, idiosyncrasies \cite{Blomsma_Vaitonyté_Skantze_Swerts_2024}, conversation topic, dialogue state, visual cues, grounding, and world knowledge. Another potential extension involves incorporating non-verbal backchannels, like facial expressions, gaze, head nods and gestures.

%The datasets used are of relatively low quality and are restricted to telephonic conversations. There can be a need for recordings of better quality and in other domains to see the full potential of context-feedback modeling.

%As backchannels often occur in places of potential turn takes~\cite{10.1007/s10849-020-09328-1}, we want to address the possibility of synthesizing such backchannels and investigate the importance of how they are spoken as well as where they occur. To do this, we need to detect transition relevance~\cite{turn} or backchannel relevance places~\cite{heldner2013backchannel}, for which existing turn-prediction models can be used~\cite{ekstedt22_interspeech}.

Another area of work can be the analysis of feedback similarity and predictability with respect to their linguistic properties and place of occurrence in the dialogue. We hypothesize that there is correlation between the confidence of feedback ranking or prediction and the different phonetic-lexical properties of the immediate context. Moreover, as mentioned earlier, the prosody of dialogue markers can depend on the future context~\cite{nath22_speechprosody}, which can also be taken into account in these analyses.

\section{Acknowledgements}

This work was partially supported by the Wallenberg AI, Autonomous Systems and Software Program (WASP) funded by the Knut and Alice Wallenberg Foundation, as well as Riksbankens Jubileumsfond (RJ) P20-0484, and Swedish Research Council (VR) 2020-03812. The authors would like to thank Erik Ekstedt for his helpful comments.

\bibliographystyle{IEEEtran}
\bibliography{main}

\end{document}